%% file: main_camera_ready.tex
\documentclass[letterpaper]{article} 
\usepackage{aaai2026}  
\usepackage{times}  
\usepackage{helvet}  
\usepackage{courier}  
\usepackage[hyphens]{url}  
\usepackage{graphicx} 
\usepackage{natbib}  
\usepackage{caption} 
\usepackage{algorithm}
\usepackage{algorithmic}

\usepackage{array,url,booktabs,multirow,mathtools,epstopdf,xcolor}

\usepackage{textcomp}

\usepackage{purudefs}
\usepackage{fontawesome}
\usepackage{pifont}
\newcommand{\cmark}{\ding{51}} 
\newcommand{\xmark}{\ding{55}} 

\usepackage[utf8]{inputenc}
\usepackage[T1]{fontenc}

\newcommand{\ie}{\textit{i}.\textit{e}., }
\newcommand{\eg}{\textit{e}.\textit{g}., }

\usepackage{newfloat}
\usepackage{listings}
\DeclareCaptionStyle{ruled}{labelfont=normalfont,labelsep=colon,strut=off} 
\floatstyle{ruled}
\newfloat{listing}{tb}{lst}{}
\floatname{listing}{Listing}
\title{Large Language Models Meet Extreme Multi-label Classification: Scaling and Multi-modal Framework}
\author{
    Diego Ortego\textsuperscript{\rm 1}, Marlon Rodr\'iguez\textsuperscript{\rm 2}, Mario Almagro\textsuperscript{\rm 1}, Kunal Dahiya\textsuperscript{\rm 3},\\
    David Jim\'enez\textsuperscript{\rm 1}, Juan C. SanMiguel\textsuperscript{\rm 2}
}
\affiliations{
    \textsuperscript{\rm 1}NielsenIQ, 
    \textsuperscript{\rm 2}Universidad Autónoma de Madrid (UAM),
    \textsuperscript{\rm 3}IIT Delhi\\

}

\usepackage{bibentry}

\begin{document}

\maketitle

\input{abstract}
\input{sec1_introduction}
\input{sec2_previous_works}

\input{sec3_method}

\input{sec4_experiments}

\input{sec5_conclusion}
\input{ack}

\bibliography{aaai2026}
\clearpage
\appendix
\input{supp_material}

\end{document}

%% file: abstract.tex
\begin{abstract}
Foundation models have revolutionized artificial intelligence across numerous domains, yet their transformative potential remains largely untapped in Extreme Multi-label Classification (XMC). Queries in XMC are associated with relevant labels from extremely large label spaces, where it is critical to strike a balance between efficiency and performance. Therefore, many recent approaches efficiently pose XMC as a maximum inner product search between embeddings learned from small encoder-only transformer architectures.
In this paper, we address two important aspects in XMC: how to effectively harness larger decoder-only models, and how to exploit visual information while maintaining computational efficiency. We demonstrate that both play a critical role in XMC separately and can be combined for improved performance.
We show that a few billion-size decoder can deliver substantial improvements while keeping computational overhead manageable.
Furthermore, our Vision-enhanced eXtreme Multi-label Learning framework (ViXML) efficiently integrates foundation vision models by pooling a single embedding per image. This limits computational growth while unlocking multi-modal capabilities. Remarkably, ViXML with small encoders outperforms text-only decoder in most cases, showing that an image is worth billions of parameters.
Finally, we present an extension of existing text-only datasets to exploit visual metadata and make them available for future benchmarking.
Comprehensive experiments across four public text-only datasets and their corresponding image enhanced versions validate our proposals' effectiveness, surpassing previous state-of-the-art by up to +8.21\% in P@1 on the largest dataset.
ViXML's code is available at: \url{https://github.com/DiegoOrtego/vixml}
\end{abstract}

%% file: sec1_introduction.tex
\section{Introduction}
\label{sec1:intro}

Scaling neural network architectures has proven to be an effective strategy for improving the performance of baseline models, as evidenced by powerful foundation models across natural language processing~\cite{Brwon20,Dubey24,Abdin24,Yang25} and computer vision~\cite{Radford21,Zhai23,Tschannen25,Bolya25}. Despite the success of this scaling paradigm, efficiency and computational constraints pose significant challenges when leveraging Large Language Models (LLMs) for complex real-world tasks. In particular, this paper tackles this challenge for eXtreme Multi-label Classification (XMC)~\cite{Babbar17,Zhang21b,Dahiya21,Jain23,Gupta24,Prabhu25}, where the task is to predict the most relevant subset of labels for a given query from an extremely large label space, often containing millions of labels~\cite{XMLRepo}, \eg in product recommendation, sponsored search and document tagging applications.

\begin{figure}[t!]
\centering{}\includegraphics[width=1.0\columnwidth]{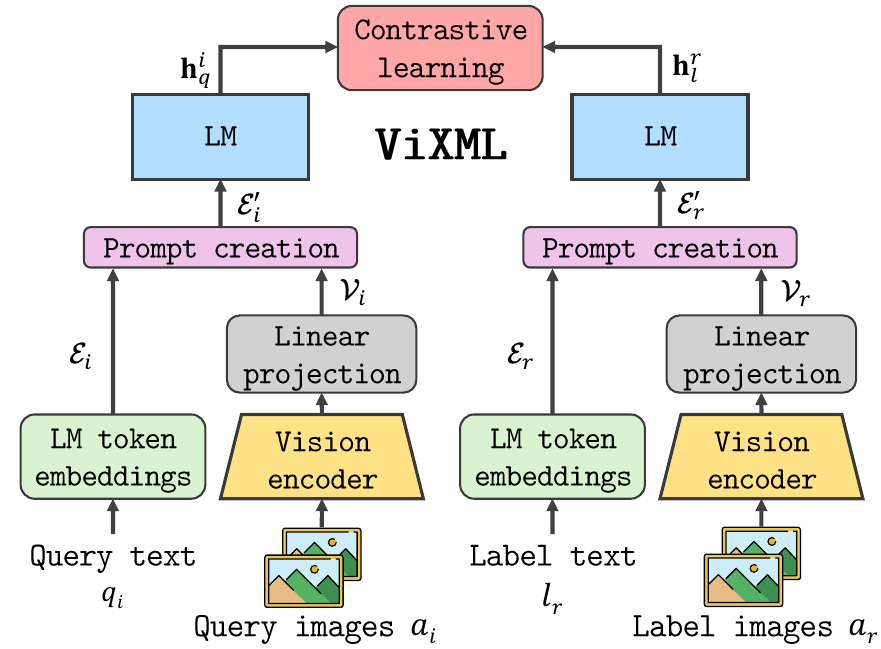}
\caption{Overview of ViXML multi-modal framework, which supports both encoder language models (LMs) and decoder LLMs. ViXML efficiently incorporates visual metadata in queries ($a_i$) and labels ($a_r$) while freezing the vision encoder for efficiency. Prompts ($\cE'_{i}$ and $\cE'_{r}$) combine text and projected image embeddings ($\cE$ and $\cV$). Sentence embeddings ($\vh^{i}_{q}$ and $\vh^{r}_{l}$) are learned via contrastive learning.
}
\label{fig:ViXML}
\end{figure}

Balancing high-quality predictions with computational efficiency is paramount for XMC tasks. Despite the success of decoder-only LLMs in providing better quality embeddings~\cite{Lee2025,Zhang25}, their adoption in XMC is still an open challenge.
While this scaling paradigm represents one potential avenue for performance enhancement, it introduces substantial computational overhead. Therefore, we also study efficient strategies to boost XMC performance. In particular, we devote our attention to item metadata, which has shown improvements in recent work~\cite{Mohan24,Prabhu25} by exploiting category or hyperlink information. However, limited attention has been directed towards exploiting visual metadata, with MUFIN~\cite{Mittal22} being the only notable example to date.

To address these challenges, this paper embraces dual-encoder learning~\cite{Dahiya21b,Gupta24} for XMC, also known as Siamese-style training. This approach trains transformer encoders using contrastive learning, addressing XMC via maximum inner product search of the extracted embeddings. Building upon this learning paradigm, we explore scalability and efficiency in two ways: (1) we propose a simple yet effective strategy for the adoption of decoder-only transformers, scaling them up to 7B parameters, which strikes a balance between capacity and efficiency for XMC; (2) we explore images as a key metadata and design an efficient approach to inject visual information using foundation vision models. 
For decoder-only models, we embed texts within structured prompt templates that provide the LLM with contextual information about the input, and extract sentence embeddings to perform dual-decoder learning. Additionally, we introduce Vision-enhanced eXtreme Multi-label Learning (ViXML), a general multi-modal framework that can be paired with any Siamese-style method for efficiently incorporating visual metadata into XMC. ViXML utilizes a single image embedding per image and concatenates it with input text token embeddings.
For decoder models, we modify the structured prompt template to explicitly acknowledge the availability of image information. Crucially, ViXML preserves efficiency by eliminating the need to train visual encoders and avoiding substantial increases in sequence length.
In Figure~\ref{fig:ViXML}, we present a general diagram of ViXML. Our main contributions are as follows:
\begin{itemize}
    \item We propose a dual-decoder learning approach to effectively adapt decoder-only architectures for XMC.
    \item We introduce ViXML, a novel multi-modal framework that incorporates images as highly effective metadata for Siamese-style XMC.
    \item We extend three text-only datasets with visual metadata from Amazon Reviews~\cite{Hou24} and make them publicly available to foster multi-modal XMC research.
    \item We systematically examine how scaling transformer backbones affects performance, achieving state-of-the-art results with the biggest decoder-only models.
    \item We comprehensively evaluate ViXML in several datasets against state-of-the-art methods, reaching remarkable improvements that range from +5.07\% to +8.21\% points in P@1. Notably, ViXML with a 66M parameter encoder outperforms text-only billion-parameter models, demonstrating the effectiveness of visual metadata.

\end{itemize}

%% file: sec2_previous_works.tex
\section{Related work}
\label{sec2:prev_work}

\paragraph{Extreme Multi-label Classification (XMC).} Siamese-style XMC using contrastive learning has received significant attention in recent years~\cite{Kharbanda24,Gupta24,Mohan24,Dahiya25,Prabhu25} due to their efficiency in handling a large number of labels.
Early contrastive learning approaches~\cite{Dahiya21b} did not use deep neural networks and their subsequent adoption marked a significant advancement in model performance \cite{Dahiya23b,Gupta24,Dahiya25}. For example,~\cite{Dahiya23} reduced training complexity with negative sampling, while~\cite{Gupta24} demonstrated state-of-the-art results using all negative labels batch-wise.
To overcome short-text ambiguity~\cite{Dahiya23b,Dahiya25} proposed enriched label representations.
Alternatively, embedding-based predictions can be improved by learning extreme classifiers~\cite{Dahiya23} or jointly training the embeddings and classifiers~\cite{Kharbanda24}.
Other works pose XMC as meta-classifier learning~\cite{Jiang21,Mittal21b,Chien23,Zhang21b} or brute-force extreme classifier learning~\cite{Jain23}.

\textbf{Metadata in XMC.}
Short-text ambiguity in queries and labels limits XMC performance. In~\cite{Mittal22} they demonstrated that visual cues can boost performance, while authors in~\cite{Mohan24} exploited textual metadata (e.g. category information or hyperlinks) available only at training time and inferred it at test time. More recently, \cite{Prabhu25} also assumed availability during training and exploited metadata to learn a teacher model that is later leveraged during metadata-free training.
In this paper, we extend existing XMC Amazon datasets with images and exploit them as metadata for training and inference.

\textbf{Text embeddings with decoder-only LLMs.} The text-embedding community has successfully adopted LLMs. For example, authors in~\cite{Jiang24} demonstrated that prompting an LLM with “\textit{This sentence: [text] means in one word:}" provides a robust text representation in the last token. More recently, authors in~\cite{Thirukovalluru25} extended this idea by ensembling representations for the same text with different perturbations. 
Nevertheless, LLM finetuning improves results~\cite{Jiang24,Behnamghader24} and numerous works followed this path studying: uni-directional vs bi-directional attention~\cite{Behnamghader24,Lee2025,Zhao25}, in-context-learning~\cite{Li25} or keeping generative capabilities~\cite{Muennighoff25}. Task adaptation is usually conducted with query instructions~\cite{Lee2025,Zhang25}.
Recent embedding-based XMC~\cite{Dahiya23b, Gupta24} predominantly use deep encoders. However, to the best of our knowledge, only two works exploited LLMs with no clear gains: QUEST~\cite{Zhou24} and MOGIC~\cite{Prabhu25}. QUEST results with Llama-7B significantly underperformed state-of-the-art encoder models, while MOGIC's LLM based oracles failed to surpass encoder-only ones. 
Therefore, effectively leveraging decoder-only LLMs for Siamese-style XMC is an open challenge that we address in this work.

\textbf{Multi-modal embeddings with Vision-Language Models (VLMs).} Following the success of LLMs for text embeddings, authors in~\cite{Jiang25} recently proposed a contrastive learning framework to convert any VLM into an embedding model that addresses multiple embedding tasks using instructions. Other works embraced the same spirit: ~\cite{Liu25} employed a language-only pretraining and afterwards performed multi-modal instruction tuning to progressively enhance retrieval performance; ~\cite{Chen25} first employed bi-directional attention to enhance context-aware reasoning during continual pre-training, and later performed contrastive finetuning on diverse tasks; \cite{Gu25} performed  knowledge distillation from a powerful LLM and later followed instruction tuning with hard negatives.
In XMC, finetuning these popular VLMs would require adding hundreds of visual tokens, dramatically increasing computational requirements. We, therefore, explore how to efficiently meet multi-modality in this work.

%% file: sec3_method.tex
\section{Method}
\label{sec3:method}


\subsection{Problem formulation}
In XMC, we consider the availability of a dataset~$\cD = \{x_i, \cP_i\}_{i=1}^{M}$ with $M$ data points or queries. Here, the query $x_i = \{q_i, a_i\}$ is endowed with the textual description $q_i$ and visual metadata $a_i$. $\cP_i$ and $\cN_i$ are the set of positive and negative labels for the i-th query, respectively. Note that $\cP_i \cup \cN_i = \cY$, where $\cY$ is the set of L labels. Additionally, each label $y$ is endowed with a textual description~($l_r$) and visual metadata~($a_r$)~\cite{Dahiya21b,Kharbanda24,Gupta24,Mittal22}. We define the visual metadata sets as $\cA_q=\{a_i\}_{i=1}^{M}$ and $\cA_l=\{a_r\}_{r=1}^{L}$ for queries and labels, respectively. In practice, $a_i$ and $a_r$ can contain one or multiple images and, occasionally, be empty when no visual information is available.

Predicting the positive labels for every query $x_i$ can be achieved by posing XMC as a maximum inner product search between the query and label embeddings. The training process involves learning a function $f_{\theta}: (\cX, \cY)  \rightarrow \mathbb{R}^{d}$, where $\theta$ denotes the parameters of a neural network that separately encodes query and labels into the $d$-dimensional sentence embeddings $\vh^{i}_{q}$ and $\vh^{r}_{l}$. Note that $\vh^{r}_{l}$ can be defined as $\vh^{r}_{p}$ and $\vh^{r}_{n}$ to denote positive and negative label embeddings, respectively. The text-only setup ignores visual metadata and is a special case of our proposed formulation, which is widely used in the existing literature~\cite{Dahiya21b,Gupta24,Mohan24}.

\subsection{Dual-decoder learning}
Siamese-style XMC~\cite{Dahiya23b, Kharbanda24, Gupta24} predominantly uses deep encoders as backbone neural networks, \ie dual-encoder learning, and demonstrate that bigger backbones bring consistent improvements in performance.
From this observation, the natural step is to explore bigger decoder-only architectures and exploit them for XMC. However, as discussed in Section~\ref{sec2:prev_work}, no Siamese-style method has effectively leveraged decoder-only LLMs in this field.
These results contrast decoder LLMs' dominance in text embeddings benchmarks~\cite{Wang24,Muennighoff25,Lee2025}.
Motivated by this disparity, we investigate effective strategies to leveraging the strong generalization capabilities of LLMs and propose a text-only dual-decoder learning methodology that boosts encoder-only performance while maintaining manageable computational requirements.

\textbf{Prompting.} We define a sequence of input embeddings $\cE'_{i}$ for each query $q_i$, so queries are embedded within a structured prompt template. We do so by concatenating to the sequence of text token embeddings $\cE_{i}=\{\ve_0, \ve_1, \ve_2,...,\ve_E\}$, a prefix, and end-of-sequence (EOS) token as
\begin{equation}
\cE'_{i} = \cT~\oplus~\cE_{i}~\oplus~\ve_{EOS},
\end{equation}
where $\oplus$ denotes the concatenation operator and $\cT$ is the sequence of token embeddings for the prefix. In practice, we use the text prefix ``\textit{This product text}" and the end-of-sequence  ``\textit{\textless\textbar endoftext\textbar\textgreater}" with token embedding $\ve_{EOS}$. The same structured prompt is applied to each label $l_r$ to build $\cE'_{r}$ from its sequence of text token embeddings $\cE_{r}$. This concise template enables input contextualization and keeps a low sequence length to constrain the memory overhead.
Alternative prefix and EOS choices are analyzed in Section~\ref{sec:prompting_effect}.

\textbf{Embedding extraction.} During LLMs forward pass we keep uni-directional attention as we found it to work well and aligns with pre-training. However, we leave for future work the open question of whether bi-directional~\cite{Behnamghader24,Lee2025} or uni-directional~\cite{Zhao25} attention works best. 

\textbf{Optimization.} The proposed dual-decoder learning framework can exploit any contrastive learning strategy. In this paper we adopt the triplet loss as proposed in \cite{Dahiya23} for either encoder or decoder architectures:
\begin{equation}
\cL=\sum_{i=1}^{B}\underset{\substack{j\in\mathcal{P}_i\\
k\in\mathcal{N}_i
}
}{\sum}[\vh_{q}^{i}\cdot \vh_{n}^{k}-\vh_{q}^{i}\cdot \vh_{p}^{j}+m]_{+} ,
\label{Eq:TripletLoss}
\end{equation}
where $B$ denotes the number of batch queries, $m$ is a margin and $\vh$ refers to L2-normalized embeddings. For efficiency, we use NGAME~\cite{Dahiya23} hard negative mining strategy, thus introducing abuse of notation for $\cN_i$ when using it to denote the subset of negatives selected rather than all negative labels. This vanilla optimization is used in NGAME~\cite{Dahiya23}, however there are better ways of learning embeddings for XMC~\cite{Kharbanda24,Mohan24,Dahiya25}. In the first half of Table~\ref{tab:results_mm_ngame_dexa} (Section~\ref{sec:results_mm}) we compare the performance of three approaches (NGAME, DEXA~\cite{Dahiya23b} and PRIME~\cite{Dahiya25}). In the remaining of the paper and unless otherwise stated, we adopt PRIME as our base method. In particular, PRIME introduces a shallow label prototype network to aggregate text embeddings, learnable auxiliary vectors and centroids for building enriched label representations. Note that PRIME extends the optimization in Eq.~\ref{Eq:TripletLoss} with additional terms and we refer the reader to \cite{Dahiya25} for a detailed explanation.

Fine-tuning LLMs is resource-intensive, and inference often suffers from high latency compared to encoder-only transformers. To mitigate these challenges, we adopt relatively small LLMs (up to 7B parameters) and leverage their sample efficiency capabilities to reduce training times. While inference throughput will always be impacted due to its increased parameter size, specialized hardware or kernel-based implementations like vLLM, can make small LLMs practical for XMC.
The results of text-only dual-decoder learning are presented in Section~\ref{sec:results_dec_text}, showing that scaling to decoder-only LLMs clearly surpass encoder architectures.


\subsection{The ViXML framework}

The adoption of item metadata~\cite{Mittal22,Mohan24,Prabhu25} poses a promising path towards more robust XMC. Following this trend, we propose Vision-enhanced eXtreme Multi-label Learning (ViXML), a multi-modal framework that extends Siamese-style XMC to leverage visual metadata. ViXML is architecture-agnostic, supporting both encoder and decoder models, and operates by integrating image embeddings $\vv$ together with text token embeddings $\ve$. In particular, we use a pre-trained foundation vision model $g_{\phi}: \cA \rightarrow \mathbb{R}^{m}$ with frozen parameters $\phi$ to map images into $m-$dimensional image embeddings.
 We follow works like~\cite{Liu2023} and learn a linear layer $w_{\psi}: \mathbb{R}^{m} \to \mathbb{R}^{d}$ with parameters $\psi$ for adaptation of visual information to text inputs. We leave the study of more complex adaptation choices for future work.
ViXML is, therefore,  computationally efficient as: (i) utilizing a single embedding per image preserves low sequence lengths, and (ii) employing a frozen vision encoder enables storing its embeddings as a feature bank, thereby minimizing the memory overhead during training. We detail below ViXML for both encoder and decoder architectures.

\subsubsection{ViXML with encoder models}
Our framework can be seamlessly integrated into encoder models with minimal computational overhead. We do so by concatenating the sequence of image embeddings $\cV_{i}=\{\vv_0, \vv_1, \vv_2,...,\vv_{V_{i}}\}$ for the $i^{th}$ query with its corresponding sequence of text token embeddings $\cE_{i}$. Then, the final sequence of input embeddings for query $q_i$ is
\begin{equation}
\cE'_{i} = \cV_{i}~\oplus~\cE_{i}.
\end{equation}
Similar concatenation builds the label sequence $\cE'_{r}$. During forward propagation through the network, text and visual representations are enriched with each other through attention mechanisms, leading to multi-modal sentence embeddings $\vh_q^i$ and $\vh_l^r$ for queries and labels, respectively.

\subsubsection{ViXML with decoder models}
As presented for text-only dual-decoder learning, we embed images and text within a structured prompt template, which can be defined as
\begin{equation}
\cE'_{i} = \cT~\oplus~\cE_{i}~\oplus~\cI~\oplus~\cV_{i}~\oplus~\ve_{EOS},
\end{equation}
where $\cT$ and $\cI$ are the sequences of token embeddings for the text and image prefixes, respectively. We build multi-modal sentence embeddings $\vh_q^i$ and $\vh_l^r$ by mean pooling token representations.
We initially mimicked the encoder mechanism to build the input sequence $\cE'_{i}$ and fell into two main problems: (i) a drop in performance compared to using a structured prompt template, and (ii) degenerated performance when concatenating image embeddings as the first token in the sequence. Prompting and placing image information after the text, both helped in achieving robust performance (see Section~\ref{sec:prompting_effect}). Note that the latter decision implies contextualizing image information with text given the uni-directional attention of decoder LLMs.

%% file: sec4_experiments.tex
\section{Experimental work}
\label{sec4:experiments}

\subsection{Framework}
\paragraph{Details.} We experiment with four publicly available XMC text-only datasets~\cite{XMLRepo} for product-to-product recommendation; only MM-AmazonTitles-300K has visual metadata available. Therefore, for multi-modal evaluation we extend LF-AmazonTitles-131K, LF-Amazon-131K and LF-AmazonTitles-1.3M, for which images can be extracted from Amazon Reviews data~\cite{Hou24}. In particular, we selected image URLs from 2023 version and moved towards older versions if no images were available\footnote{URLs \& embeddings: \url{https://github.com/DiegoOrtego/vixml}.}.
Note that these datasets vary the amount of labels and text length, thus supporting a robust evaluation in diverse contexts. We present dataset details in the supplementary material.
For evaluation, we adopt standard XMC metrics defined at~\cite{XMLRepo}, \textit{e.g.} precision, propensity-scored precision or recall (P@$k$, PSP@$k$ and R@$k$).

\paragraph{Baseline methods.}
We compare against the following text-only methods (extended comparison in the supplementary material): DEXA~\cite{Dahiya23b}, PINA~\cite{Chien23}, Renée~\cite{Jain23}, DEXML~\cite{Gupta24} UniDEC~\cite{Kharbanda24}, OAK~\cite{Mohan24}, PRIME~\cite{Dahiya25} and MOGIC~\cite{Prabhu25}. We further compare with MUFIN~\cite{Mittal22} on MM-AmazonTitles-300K.

\paragraph{Implementation details.} We train encoder models of different sizes: MiniLM-L3, DistilBERT and BERT~\cite{Reimers19, Devlin19}. For decoder models, we focus on Qwen2.5-Instruct~\cite{Yang25}, but experiment with other LLMs in Section~\ref{sec:results_dec_text} to demonstrate generalization.
Unless otherwise stated, we adopt SigLIPv2~\cite{Tschannen25} as vision encoder and aggregate tokens with model's multi-head attention pooling, use 3 images in ViXML and perform mean pooling for sentence embedding extraction.
We use low-rank adaptation (LoRA) during finetuning of decoder models and run all experiments in a single 80GB GPU as the budget constraint.
We always report performance in the last training epoch, \ie we do not conduct any custom checkpoint selection based on validation metrics. 
We provide further implementation details, inference latencies, a study on the impact of image count and an example of random seed variability (standard deviation of 0.103 in P@1) in the supplementary material.

\begin{figure}[t!]
\centering{}\includegraphics[width=1.0\columnwidth]{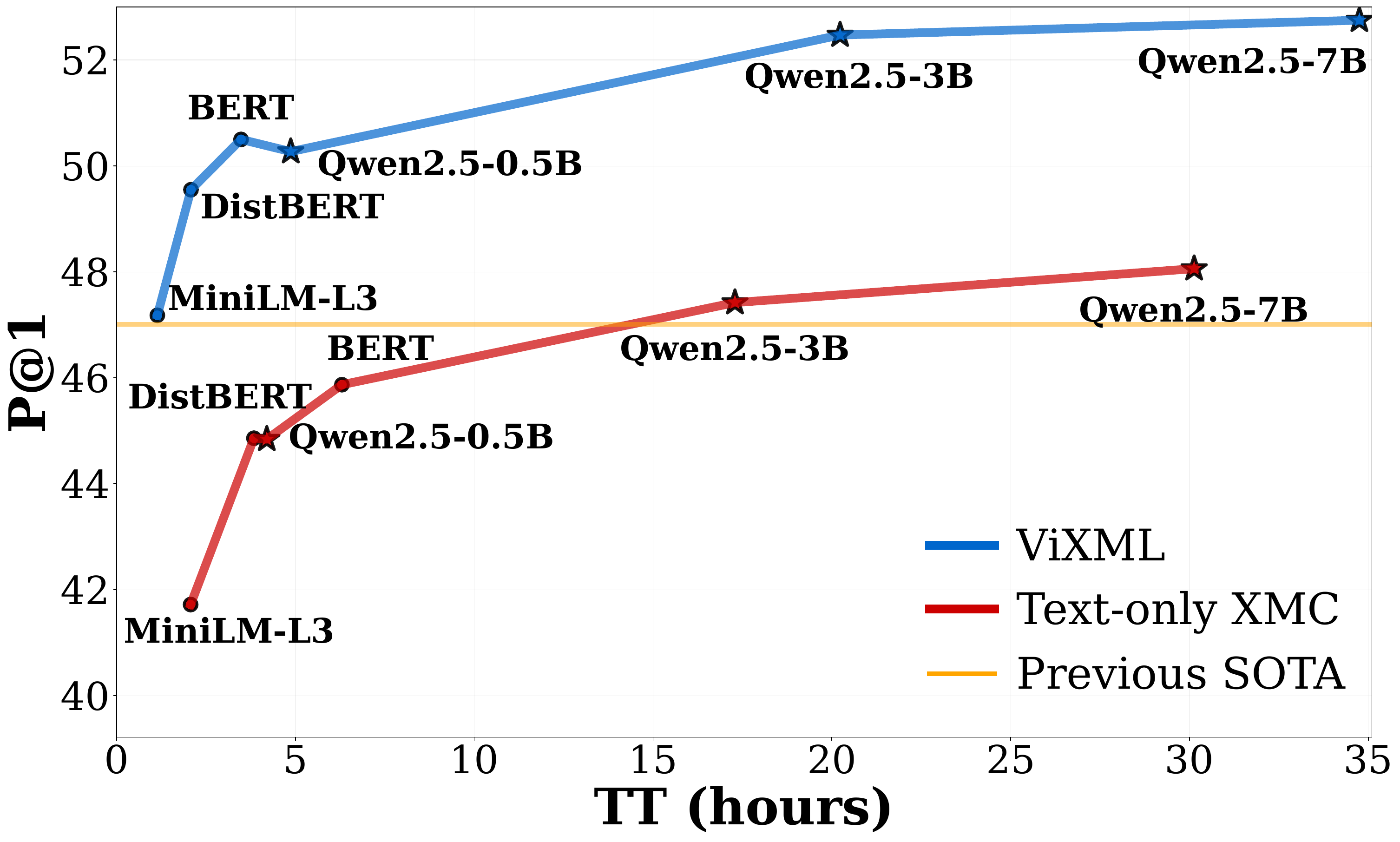}
\caption{Performance (P@1) and Training Time (TT) for dual-encoder (dots) and dual-decoder (stars) learning in LF-AmazonTitles-131K. The ViXML multi-modal framework (blue) improves text-only alternatives (red), while decoder models boost encoder performance in both setups. Previous state-of-the-art (SOTA) is represented by MOGIC method.}
\label{fig:TT}
\end{figure}

\input{results_text}
\input{results_text_LLMfamilies}

\subsection{Scaling text-only architectures for XMC} \label{sec:results_dec_text}
Related literature mostly operates with DistilBERT 66M parameters model which keeps XMC extremely efficient, but hinders understanding the scaling potential. We, therefore, scale in this work encoder sizes and, more importantly, propose a dual-decoder learning strategy to leverage decoder-only LLMs. We adopt Qwen2.5-Instruct with varying sizes, reaching 3B parameters for all datasets and 7B in LF-AmazonTitles-131K. Table~\ref{tab:results_scaling_text} presents our results (we discard visual metadata for MM-AmazonTitles-300K), demonstrating that foundation LLMs can deliver better performance for XMC.
Additionally, we keep computational overhead manageable as (i) all our experiments fit on a single GPU, and (ii) we bound the ineludible training time growth when scaling to decoder-only models by significantly reducing the number of training epochs. Red line in Figure~\ref{fig:TT} shows the performance and training times across backbone architectures in LF-AmazonTitles-131K. We demonstrate that scaling the size of the models from encoder to decoder-only architectures surpasses previous state-of-the-art results using text-only inputs (orange line). Despite requiring longer training times, decoder-only models are more sample efficient; encoders are trained for 300 epochs and decoders only for 30, which effectively reduces training times by lowering the number of training epochs one order of magnitude. Note that 0.5B decoder model and 66M distilBERT train in comparable time in our setup.

\paragraph{Different families of LLM.} We demonstrate in the second half of Table~\ref{tab:results_LLMzoo} that text-only dual-decoder learning with Llama-3.2, Gemma-3 and Qwen variants achieve robust results in LF-AmazonTitles-131K, surpassing encoder models reported in Table~\ref{tab:results_scaling_text}.
While we focus on task-specific XMC embeddings, strong general-purpose embeddings might appear adequate. However, Table~\ref{tab:results_LLMzoo} shows that the off-the-shelf Qwen3-Embedding-4B model performs significantly worse than all trained alternatives. Additionally, we tested Qwen3-Emb as recommended in \cite{Zhang25} (*), \ie following a prompt template for the queries (``\textit{Instruct: Retrieve a semantically related product.\textbackslash nQuery: [text] \textless\textbar endoftext\textbar\textgreater}") and not for the labels, while extracting embeddings with last token pooling. As reported, this strategy did not lead to improvements.

\input{results_mm}

\subsection{Exploiting visual meta-data}
\label{sec:results_mm}
This section presents the benefits of ViXML multi-modal framework to leverage visual metadata in XMC. First, Table~\ref{tab:results_mm} presents ViXML's performance when paired with the same backbones of Table~\ref{tab:results_scaling_text}. Notably, a small DistilBERT model with 66M parameters paired with ViXML surpasses the best performance achieved with billion-size text-only models in most cases.
These results underscore that an image is worth billions of parameters for achieving robust and efficient XMC.
There is only one case where encoder-based ViXML approaches do not surpass text-only 3B model (LF-Amazon-131K). In contrast to the rest of the datasets, LF-Amazon-131K contains product titles and descriptions. We argue that decoder-only models benefit from longer textual inputs by exploiting their parametric knowledge, struggling in those cases where only product titles are presented.

Regarding training times, Figure~\ref{fig:TT} presents the impact of ViXML (blue line) over text-only training (red line) in LF-AmazonTitles-131K. Remarkably, ViXML with encoder models significantly reduces training time due to a faster convergence to top performance that enables halving the training epochs (300 to 150). On the other hand, for decoder models we only train for 30 epochs for both text-only and ViXML models, showing that ViXML introduces an overhead of around 15\%-17\% which mostly comes from longer inputs due to prompts and visual metadata.

\paragraph{Early-fusion or late-fusion.} ViXML framework conducts early-fusion of visual metadata, while MUFIN employs late-fusion via self-attention to integrate text and image embeddings. To demonstrate the benefits of our early-fusion strategy, we compare ViXML and MUFIN in Table~\ref{tab:results_mufin}.
Using MUFIN's ViT-32 encoder, ViXML achieves +1\% in P@1 (53.30 vs 52.30) despite requiring only an encoder training (MUFIN trains extreme classifiers and prediction combination). However, this comparison has limitations due to differences in training stages, image counts, and Siamese-style methods. For a fair comparison, we pair both MUFIN and ViXML with the same XMC method to isolate the impact of visual metadata injection. This evaluation reveals that the proposed early fusion consistently outperforms MUFIN over 1.5\% in P@1 across both ViT-32 and SigLIP2 encoders. Additionally, these results highlight the importance of leveraging large foundation vision models, as performance significantly improves when scaling from 86M ViT-32 to 1.14B SigLIP2 in both MUFIN and ViXML.

\input{results_mufin}


\paragraph{ViXML across methods.} We use PRIME~\cite{Dahiya25} as the base method due to its efficiency and robustness. However, ViXML is a general multi-modal framework that can be used with any Siamese-style method. We support this claim by pairing ViXML with NGAME~\cite{Dahiya23}, DEXA~\cite{Dahiya23b} and PRIME ~\cite{Dahiya25} methods in LF-AmazonTitles-131K dataset (see Table~\ref{tab:results_mm_ngame_dexa}). The results demonstrate that ViXML delivers an important boost across all metrics for all methods.
\input{results_mm_ngame_dexa}

\paragraph{Pre-trained VLMs.} Section~\ref{sec2:prev_work} discussed VLM-based embeddings and their challenging adoption in XMC due to using hundreds of visual tokens. Conversely, the proposed ViXML stands as an efficient alternative. However, a reasonable question to address is whether off-the-shelf VLMs can generalize well to XMC. Table~\ref{tab:results_LLMzoo} demonstrates that recent off-the-shelf embedding VLM MoCa~\cite{Chen25} yiels weak results as compared to finetuned alternatives.

\subsection{Prompting effect}
\label{sec:prompting_effect}
Table~\ref{tab:resuts_prompts} compares the performance of different strategies to prompting decoder-only models with and without images (Qwen2.5-0.5B-Instruct trained for 20 epochs on LF-AmazonTitles-131K dataset).
For text-only inputs, adding prefix $\cT_1$ (``\textit{This product text}") and end-of-sequence token $\ve_{EOS}$ (``\textit{\textless\textbar endoftext\textbar\textgreater}") both improve performance over not using them. With visual metadata, prefixes $\cT_1$ or $\cT_2$ (``\textit{and its text}") for text and $\cI_1$ (``\textit{This product image}") or $\cI_2$ (``\textit{and its image}") also benefit performance.
These results, together with further prompt variations reported in the supplementary material, suggest that performance gains stem from structural cues that help leveraging pretrained LLMs as no additional information is injected with these prompts.
We also investigate the order of visual tokens in the input sequence for decoder models, thus we need to account for the pretraining dynamics of these models. As demonstrated in ~\cite{Barbero25}, pretraining LLMs generates attention sinks on the first token to avoid representational collapse. We observe that phenomena when breaking the pretraining dynamics by placing image tokens first without prefixes ($\cV~\oplus~\cE~\oplus~\ve_{EOS}$), significantly dropping performance. Conversely, placing images at the end and using prefixes improves performance, thus we adopt this robust configuration across the experiments.

\input{abs}


\subsection{Comparative evaluation}
\input{results_comparison}
We conclude our experimentation in Table~\ref{tab:results_comparison}, which shows large improvements of the ViXML framework (with dual-decoder learning and PRIME) over the best previous results across all metrics and datasets, ranging from +5.07\% to +8.21\% points in P@1. Note that MOGIC does not balance PSP@1 and P@1, reporting +0.88\% in PSP@1 with respect to ViXML in LF-AmazonTitles-1.3M, while dropping 16.88\% in P@1.
As many other works, we incorporate inference enhancements for ViXML. In particular, we extend the search space using training data samples to refine label predictions~\cite{Wang25,Dahiya21}. We refer the reader to the supplementary material for further details.

%% file: results_text.tex
\begin{table}[t]
      \centering
      \resizebox{0.95\linewidth}{!}{
        \begin{tabular}{@{}l|cccc@{}}
        \toprule
        \textbf{Backbone} & \textbf{P@1} & \textbf{P@5} & \textbf{PSP@1} & \textbf{R@100}  \\ \midrule

        \multicolumn{5}{c}{\textbf{LF-AmazonTitles-131K}}\\ \midrule
        \textbf{MiniLM-L3} & 41.72 & 20.16 & 36.17 & 64.68 \\
        \textbf{DistilBERT} & 44.86 & 21.45 & 39.62 & 68.05 \\
        \textbf{BERT} & 45.87 & 21.88 & 40.54 & 69.67 \\
        \midrule
        \textbf{Qwen2.5-0.5B-I} & 44.84 & 21.66 & 39.64 & 70.30 \\
        \textbf{Qwen2.5-3B-I} & 47.42 & 22.89 & 41.88 & 74.34 \\
        \textbf{Qwen2.5-7B-I} & \textbf{48.06} & \textbf{23.07} & \textbf{42.43} & \textbf{75.24} \\
        \midrule
        \multicolumn{5}{c}{\textbf{MM-AmazonTitles-300K}}\\ \midrule
        \textbf{MiniLM-L3} & 49.54 & 31.93 & 33.83 & 69.30 \\
        \textbf{DistilBERT} & 52.40 & 34.29 & 36.36 & 72.59 \\
        \textbf{BERT} & 52.95 & 34.57 & \textbf{36.85} & 73.34 \\
        \midrule
        \textbf{Qwen2.5-0.5B-I} & 52.30 & 34.23 & 36.22 & 74.44 \\
        \textbf{Qwen2.5-3B-I} & \textbf{54.71} & \textbf{35.93} & 36.29 & \textbf{77.22} \\
        \midrule
        \multicolumn{5}{c}{\textbf{LF-AmazonTitles-1.3M}}\\ \midrule
        \textbf{MiniLM-L3} & 51.59 & 39.28 & 29.50 & 58.44 \\
        \textbf{DistilBERT} & 58.49 & 45.27 & 33.28 & 63.76 \\
        \textbf{BERT} & 59.10 & 45.62 & \textbf{34.10} & 64.30 \\
        \midrule
        \textbf{Qwen2.5-0.5B-I} & 58.02 & 45.11 & 31.68 & 63.96 \\
        \textbf{Qwen2.5-3B-I} & \textbf{60.74} & \textbf{47.39} & 34.01 & \textbf{66.88} \\
        \midrule
        \multicolumn{5}{c}{\textbf{LF-Amazon-131K}}\\ \midrule
        \textbf{MiniLM-L3} & 41.99 & 20.42 & 35.44 & 68.37 \\
        \textbf{DistilBERT} & 47.98 & 23.21 & 40.99 & 76.22 \\
        \textbf{BERT} & 49.15 & 23.71 & 42.18 & 78.94 \\
        \midrule
        \textbf{Qwen2.5-0.5B-I} & 49.23 & 23.99 & 42.40 & 81.24 \\
        \textbf{Qwen2.5-3B-I} & \textbf{53.17} & \textbf{25.72} & \textbf{45.74} & \textbf{85.61} \\
        \bottomrule
    \end{tabular}
    }
    \caption{Impact of backbone size on text-only datasets.}
    \label{tab:results_scaling_text}
\end{table}

%% file: results_text_LLMfamilies.tex
\begin{table}[t!]
      \centering
      \resizebox{0.97\linewidth}{!}{
        \begin{tabular}{@{}l|l|cccc@{}}
        \toprule
        \textbf{Backbone} & \textbf{\#p} & \textbf{P@1} & \textbf{PSP@1} & \textbf{R@100}  \\
        \midrule
        \textbf{MoCa*~\faLock} & 3.75B & \textbf{25.67} & \textbf{26.31} & \textbf{54.48} \\
        \textbf{Qwen3-Emb*~\faLock} & 4.02B & 18.33 & 18.32 & 42.67 \\
        \textbf{Qwen3-Emb~\faLock} & 4.02B & 22.33 & 22.49 & 46.69 \\
        \midrule
        \textbf{Qwen2.5-I} & 3.09B & 47.42 & 41.88 & \textbf{74.34} \\
        \textbf{Llama-3.2} & 3.21B & \textbf{48.19} & \textbf{42.60} & 73.95 \\
        \textbf{Qwen3} & 4.02B & 47.24 & 41.79 & 73.42 \\
        \textbf{Qwen3-Emb} & 4.02B & 47.51 & 42.11 & 73.62 \\
        \textbf{Gemma-3-I} & 4.3B & 47.74 & 42.34 & 73.70 \\
        
        \bottomrule
    \end{tabular}
    }
    \caption{Pre-trained general-purpose embeddings (\faLock) vs dual-decoder learning with different LLM families. \#p is the number of parameters and (*) denotes using the authors' embedding extraction methodology. MoCa is a vision-language embedding model, \ie it uses visual metadata. 
    }
    \label{tab:results_LLMzoo}
\end{table}

%% file: results_mm.tex
\begin{table}[t]
      \centering
      \resizebox{0.95\linewidth}{!}{
        \begin{tabular}{@{}l|cccc@{}}
        \toprule
        \textbf{Backbone} & \textbf{P@1} & \textbf{P@5} & \textbf{PSP@1} & \textbf{R@100}  \\ \midrule

        \multicolumn{5}{c}{\textbf{LF-AmazonTitles-131K}}\\ \midrule
        \textbf{MiniLM-L3} & 47.18 & 22.69 & 41.01 & 73.81 \\
        \textbf{DistilBERT} & 49.55 & 23.73 & 43.78 & 76.73 \\
        \textbf{BERT} & 50.50 & 24.08 & 44.67 & 78.43 \\
        \midrule
        \textbf{Qwen2.5-0.5B-I} & 50.27 & 24.22 & 44.31 & 79.41 \\
        \textbf{Qwen2.5-3B-I} & 52.47 & 25.26 & 46.46 & 82.21 \\
        \textbf{Qwen2.5-7B-I} & \textbf{52.75} & \textbf{25.37} & \textbf{46.70} & \textbf{83.27} \\
        \midrule
        \multicolumn{5}{c}{\textbf{MM-AmazonTitles-300K}}\\ \midrule
        \textbf{MiniLM-L3} & 52.97 & 34.13 & 35.51 & 73.08 \\
        \textbf{DistilBERT} & 55.03 & 35.91 & 37.52 & 76.07 \\
        \textbf{BERT} & 55.59 & 36.10 & 37.83 & 76.67 \\
        \midrule
        \textbf{Qwen2.5-0.5B-I} & 55.28 & 36.09 & 36.48 & 77.61 \\
        \textbf{Qwen2.5-3B-I} & \textbf{56.55} & \textbf{37.04} & \textbf{37.60} & \textbf{79.38} \\
        \midrule
        \multicolumn{5}{c}{\textbf{LF-AmazonTitles-1.3M}}\\ \midrule
        \textbf{MiniLM-L3} & 60.36 & 45.78 & 34.88 & 67.35 \\
        \textbf{DistilBERT} & 64.17 & 49.43 & 36.58 & 70.65 \\
        \textbf{BERT} & 64.59 & 49.66 & 37.35 & 71.08 \\
        \midrule
        \textbf{Qwen2.5-0.5B-I} & 63.57 & 49.08 & 35.21 & 70.35 \\
        \textbf{Qwen2.5-3B-I} & \textbf{66.01} & \textbf{51.21} & \textbf{37.71} & \textbf{73.13} \\
        \midrule
        \multicolumn{5}{c}{\textbf{LF-Amazon-131K}}\\ \midrule
        \textbf{MiniLM-L3} & 46.88 & 22.63 & 39.97 & 75.80 \\
        \textbf{DistilBERT} & 51.30 & 24.69 & 44.09 & 81.40 \\
        \textbf{BERT} & 52.28 & 25.12 & 45.08 & 83.40 \\
        \midrule
        \textbf{Qwen2.5-0.5B-I} & 51.76 & 25.05 & 45.08 & 84.52 \\
        \textbf{Qwen2.5-3B-I} & \textbf{55.11} & \textbf{26.46} & \textbf{47.73} & \textbf{87.73} \\
        
        \bottomrule
    \end{tabular}
    }
    \caption{Impact of backbone size on ViXML. Note that we use visual metadata across all backbones and datasets.
    }
    \label{tab:results_mm}
\end{table}

%% file: results_mufin.tex
\begin{table}[t]
      \centering
      \resizebox{1.0\linewidth}{!}{
        \begin{tabular}{@{}l|l|l|ccc@{}}
        \toprule
        \textbf{MM} & \textbf{XMC} & \textbf{V} & \textbf{P@1} & \textbf{P@5} & \textbf{R@10}  \\ \midrule
        \textbf{MUFIN} & \textbf{MUFIN} & \textbf{ViT-32} & 52.30 & 34.76 & 50.63 \\
        \midrule
        \multirow{2}{*}{\textbf{MUFIN}}
        & \multirow{2}{*}{\textbf{PRIME}} & \textbf{ViT-32} & 52.62 & 34.35 & 49.28 \\
        & & \textbf{SigLIP2} & 53.44 & 34.79 & 50.18 \\
        \midrule
        \multirow{2}{*}{\textbf{ViXML}}
        & \multirow{2}{*}{\textbf{PRIME}} & \textbf{ViT-32} & 53.30 & 34.80 & 50.19 \\
        & & \textbf{SigLIP2} & \textbf{55.03} & \textbf{35.91} & \textbf{51.98} \\
        
        \bottomrule
    \end{tabular}
    }
    \caption{Comparison of multi-modal (MM) strategies with different vision encoders (V). First row shows original MUFIN (trains encoder, classifier and fusion). Then, we present MUFIN (late-fusion) and ViXML (early-fusion) on top of PRIME method for fair comparison.}\label{tab:results_mufin}
\end{table}

%% file: results_mm_ngame_dexa.tex

\begin{table}[t]
      \centering
      \resizebox{1.0\linewidth}{!}{
        \begin{tabular}{@{}l|l|cccc@{}}
        \toprule
        \textbf{MM} & \textbf{XMC} & \textbf{P@1} & \textbf{P@5} & \textbf{PSP@1} & \textbf{R@100}  \\ \midrule
        \multirow{3}{*}{\textbf{-}}
        & \textbf{NGAME} & 42.47 & 20.44 & 37.95 & 65.43  \\
        & \textbf{DEXA} & 44.38 & 20.96 & 38.64 & 65.37  \\
        & \textbf{PRIME} & 44.86 & 21.45 & 39.62 & 68.05  \\
        \midrule
        \multirow{3}{*}{\textbf{ViXML}}
        & \textbf{NGAME} & 47.81 & 22.79 & 43.04 & 74.09  \\
        & \textbf{DEXA} & 49.48 & 23.19 & 43.21 & 73.57  \\
        & \textbf{PRIME} & \textbf{49.55} & \textbf{23.73} & \textbf{43.78} & \textbf{76.73}  \\
        
        \bottomrule
    \end{tabular}
    }
    \caption{Effect of ViXML for different XMC methods.} \label{tab:results_mm_ngame_dexa}
\end{table}

%% file: abs.tex

\begin{table}[t!]
\centering
\setlength{\tabcolsep}{5pt}
\resizebox{0.72\linewidth}{!}{
  \begin{tabular}{@{}l|cc@{}}
  \toprule
  \textbf{Prompt templates} & \textbf{P@1} \\ \midrule
    $\cE$ & 43.14 \\ 
    $\cT_1~\oplus~\cE$  & 44.15 \\ 
    $\cT_1~\oplus~\cE~\oplus~\ve_{EOS}$  & \textbf{44.66} \\ 
    \midrule
    $\cV~\oplus~\cE~\oplus~\ve_{EOS}$  & 46.12 \\ 
    $~\cE~\oplus~\cV~\oplus~\ve_{EOS}$ & 49.45 \\ 
    $\cI_1~\oplus~\cV~\oplus~\cT_2~\oplus~\cE~\oplus~\ve_{EOS}$  & 49.29 \\ 
    $\cT_1~\oplus~\cE_{i}~\oplus~\cI_2~\oplus~\cV_{i}~\oplus~\ve_{EOS}$ & \textbf{49.67} \\
    \bottomrule
\end{tabular}
}\caption{Prompting effect for dual-decoder learning. We consider sequences of embeddings for text prefixes $\cT_1$ (``\textit{This product text}") and $\cT_2$ (``\textit{and its text}"), image prefixes $\cI_1$ (``\textit{This product image}") and $\cI_2$ (``\textit{and its image}"), text $\cE$ and images $\cV$. EOS stands for end-of-sequence token.} \label{tab:resuts_prompts}
\end{table}


%% file: results_comparison.tex
\begin{table}[t!]
\centering
\resizebox{0.95\columnwidth}{!}{
    \begin{tabular}{@{}l|cc|ccc}
        \toprule
        \textbf{Method} & \textbf{E} & \textbf{M} & \textbf{P@1} & \textbf{P@5} & \textbf{PSP@1} \\ \midrule
        \multicolumn{6}{c}{\textbf{LF-AmazonTitles-131K}} \\
        \midrule
        \textbf{DEXML} & \xmark & \xmark & 42.52 & 20.64 & -\\
        \textbf{UniDEC} & \cmark & \xmark & 44.35 & 21.03 & 39.23\\
        \textbf{PRIME} & \cmark & \xmark & 45.26 & 21.48 & 39.29\\
        \textbf{Renée} & \cmark & \xmark & 46.05 & 22.04 & 39.08\\
        \textbf{DEXA} & \cmark & \xmark & 46.42 & 21.59 & 39.11\\
        \textbf{OAK} & \cmark & \cmark & 46.42 & 21.88 & 39.76\\
        \textbf{MOGIC} & \cmark & \cmark & 47.01 & 22.40 & 40.62\\
        \midrule
        \textbf{ViXML (Ours)} & \cmark & \cmark & \textbf{53.08} & \textbf{25.74} & \textbf{46.22}\\
        \midrule
        \multicolumn{6}{c}{\textbf{MM-AmazonTitles-300K}} \\
        \midrule
        \textbf{MUFIN} & \cmark & \cmark & 52.30 & 34.76 & - \\
        \midrule
        \textbf{ViXML (Ours)} & \cmark & \cmark & \textbf{57.37} & \textbf{38.08} & \textbf{36.12} \\
        \midrule
        \multicolumn{6}{c}{\textbf{LF-AmazonTitles-1.3M}} \\
        \midrule
        \textbf{OAK} & \cmark & \cmark & 49.46 & 38.61 & 34.92\\
        \textbf{MOGIC} & \cmark & \cmark & 50.95 & 39.95 & \textbf{36.28}\\
        \textbf{Renée} & \cmark & \xmark & 56.04 & 45.32 & 28.54\\
        \textbf{DEXA} & \cmark & \xmark & 56.63 & 43.90 & 29.12\\
        \textbf{UniDEC} & \cmark & \xmark & 57.41 & 45.89 & 30.10\\
        \textbf{DEXML} & \xmark & \xmark & 58.40 & 45.46 & 31.36\\
        \textbf{PRIME} & \cmark & \xmark & 59.62 & 46.75 & 31.20\\
        \midrule
        \textbf{ViXML (Ours)} & \cmark & \cmark & \textbf{67.83} & \textbf{53.72} & 35.40\\
        \midrule
        \multicolumn{6}{c}{\textbf{LF-Amazon-131K}} \\
        \midrule
        \textbf{PINA} & \cmark & \xmark & 46.76 & 23.20 & - \\
        \textbf{DEXA} & \cmark & \xmark & 47.16 & 22.42 & 38.70 \\
        \textbf{UniDEC} & \cmark & \xmark & 47.80 & 23.35 & 40.28 \\
        \textbf{Renée} & \cmark & \xmark & 48.05 & 23.26 & 40.11 \\
        \textbf{PRIME} & \cmark & \xmark & 48.20 & 23.28 & 40.16 \\
        \textbf{OAK} & \cmark & \cmark & 48.36 & 22.20 & - \\
        \textbf{MOGIC} & \cmark & \cmark & 50.05 & 23.72 & - \\
        \midrule
        \textbf{ViXML (Ours)} & \cmark & \cmark & \textbf{55.57} & \textbf{26.84} & \textbf{47.41} \\
        
        \bottomrule 
        \end{tabular}
}
\caption{Comparison of ViXML against related work regardless of inference enhancements (E) or meta-data (M).}
\label{tab:results_comparison}
\end{table}

%% file: sec5_conclusion.tex
\section{Conclusion}
\label{sec5:conclusion}
In this paper we demonstrate the impact of the scaling laws in XMC and present for the first time an efficient strategy to enable massively pretrained decoder-only transformers for XMC. We show that increasing the backbones' size improves performance, getting state-of-the-art results.
In addition, we present ViXML, an effective approach to exploit visual information in XMC. By using an early-fusion strategy and tailored prompting templates for decoder models, we bring another significant leap in XMC performance. We also open to the community three new extensions of existing datasets where we incorporate visual metadata information.
Extensive experimentation across multiple datasets and methods demonstrate all our claims, advancing state-of-the-art performance in XMC by a significant margin. In the supplementary material we discuss the main limitations of our work and suggest potential lines for future work.

%% file: ack.tex
\section*{Acknowledgements}
The authors want to thank Javier Martínez Cebrián for the fruitful discussions during the development of this work.

%% file: supp_material.tex
\onecolumn
\makeatletter
\@twocolumnfalse
\makeatother

\begin{center}
{\Large \textbf{Supplementary material for Large Language Models Meet Extreme Multi-label Classification: Scaling and Multi-modal Framework}}
\end{center}
\vspace{0.5cm}

\input{appendix}



%% file: appendix.tex



\section{Implementation details}
\label{app:implem_details}
We set configurations across all datasets (see dataset details in Table~\ref{tab:DatasetDetails}) to fit in a single 80 GB H100 GPU and reduce the amount of training epochs when using images and LLMs.
We follow~\cite{Dahiya25} and train for 300 epochs in regular dual-encoder learning, while reducing to 150 when using ViXML. For decoder-only models, we keep affordable training times by significantly reducing the number of training epochs to 30 for all datasets except for LF-AmazonTitles-1.3M, where we train for 40 epochs. We do so to leverage its rich ground-truth, which has 22.20 labels per query (see Table~\ref{tab:DatasetDetails}), thus needing longer training to visualize several times query to positive label relations during contrastive learning (the remaining datasets have a simpler grond-truth with 2.29 and 8.13 labels per-query on average). Note that it is common to finetune LLMs for few epochs given their great parameter knowledge and the use of parameter-efficient finetuning (PEFT). Indeed, we use low-rank adaptation (LoRA) as our PEFT strategy, using rank stabilized LoRA with 256 of rank and $\alpha$ (no dropout). Adapters are used on all linear projections of self-attention and MLP modules. In early experiments, we observed weak performance with small rank and alpha values and followed standard settings ($\geq$64), though tuning to find best values might help. Additionally, we use liger-kernels in decoder-only models to speed-up training and reduce memory usage. Gradient checkpointing is further adopted to reduce the memory footprint. As for ViXML, we use a maximum of three images across experiments and report in this supplementary material the impact of sweeping that number. Note that our visual inspection after adding images to text-only datasets (removing noisy images and empty URLs) did not reveal major quality issues. As for image resolutions used during the pre-training of the vision encoders adopted: 224 for ViT-32 and 384 for SigLIP2. Please, see exact HF checkpoints used for the vision and text models in Table~\ref{tab:hf_ckpt}.

\begin{table*}[h]
\begin{centering}
\resizebox{0.90\columnwidth}{!}{
\begin{tabular}{l|cccccccc}
\midrule

\textbf{Dataset}  & \textbf{\#Q train}  & \textbf{\#L} & \textbf{\#Q test} & \textbf{\#Q/L} & \textbf{\#L/Q} & \textbf{\%I/Q train} & \textbf{\%I/Q test} & \textbf{\%I/L}\tabularnewline
\midrule

\textbf{LF-AmazonTitles-131K} & 294.8K & 131.1K & 134.8K & 5.15 & 2.29 & 79.45 & 78.50 & 95.72 \tabularnewline

\textbf{LF-Amazon-131K} & 294.8K & 131.1K & 134.8K & 5.15 & 2.29 & 79.45 & 78.50 & 95.72\tabularnewline

\textbf{MM-AmazonTitles-300K} & 586.8K & 303.3K & 260.5K & 15.73 & 8.13 & 99.58 & 99.92 & 99.59\tabularnewline

\textbf{LF-AmazonTitles-1.3M} & 2.2M & 1.3M & 0.97M & 38.24 & 22.20 & 99.01 & 99.00 & 99.65\tabularnewline
\midrule

\end{tabular}
}
\par\end{centering}
\caption{\label{tab:DatasetDetails}Dataset statistics for benchmark datasets. Key: \#Q (number of queries), \#L (number of labels), \#Q/L (number of queries per label), \#L/Q (number of labels per query), \%I/Q train (Percentage of train queries with images), \%I/Q test (Percentage of test queries with images), \%I/L (Percentage of labels with images).}

\end{table*}

We use default training hyper-parameters for PRIME as described in~\cite{Dahiya25} and introduce some minor modifications in the method for a better performance and training trade-off:
\begin{itemize}
    \item Label centroids are not updated using an exponential moving average. Instead, we update them in-between epochs using the query embeddings that are already stored for the clustering conducted during NGAME negative mining. The update conducts the average of all query embeddings related to the label at hand.
    \item We increase the number of attention heads from 1 to 8 in the label prototype network, as it introduces a negligible overhead and it was reported in~\cite{Dahiya25} to achieve better performance.
    \item We do not use the multi-positive training proposed in \cite{Dahiya25}, as it increases training time in around 40\%. Please, note that our PRIME baseline without multi-positive, but incorporating the first two modifications, reaches 58.49\% of P@1 in LF-AmazonTitles-1.3M. This result is really close to the 58.58\% reported in~\cite{Dahiya25} with multi-positive training, while training much faster.
\end{itemize}
We always report model performance in the last training epoch, \ie we do not conduct any custom checkpoint selection based on validation metrics. Note that we compute and save the final label embeddings once the training is finished and, at test time, we obtain the embeddings for test queries and conduct a maximum inner product search to compute XMC predictions.

\begin{table*}[h!]
\begin{centering}
\resizebox{0.85\columnwidth}{!}{
\begin{tabular}{l|l|l|l}
\midrule

\textbf{Model} & \textbf{\#param} & \textbf{Learning rate} & \textbf{Huggingface checkpoint}\tabularnewline
\midrule

\textbf{MiniLM-L3} & 17.4M & 5e-4 & sentence-transformers/paraphrase-MiniLM-L3-v2 \tabularnewline

\textbf{DistilBERT} & 66.4M & 2e-4 & sentence-transformers/msmarco-distilbert-base-v4\tabularnewline

\textbf{BERT} & 109M & 1e-4 & intfloat/e5-base-v2 \tabularnewline

\textbf{Qwen2.5-I} & 494M & 5e-5 & Qwen/Qwen2.5-0.5B-Instruct \tabularnewline

\textbf{Qwen2.5-I} & 3.09B & 5e-5 & Qwen/Qwen2.5-3B-Instruct \tabularnewline

\textbf{Llama-3.2} & 3.21B & 5e-5 & meta-llama/Llama-3.2-3B-Instruct \tabularnewline

\textbf{Qwen3} & 4.02B & 5e-5 & Qwen/Qwen3-4B \tabularnewline

\textbf{Qwen3-Emb} & 4.02B & 5e-5 & Qwen/Qwen3-Embedding-4B \tabularnewline

\textbf{Gemma-3-I} & 4.3B & 5e-5 & google/gemma-3-4b-it \tabularnewline

\textbf{Qwen2.5-I} & 7.62B & 5e-5 & Qwen/Qwen2.5-7B-Instruct \tabularnewline

\midrule
\textbf{ViT-32} & 86.4M & - & google/vit-base-patch16-224-in21k \tabularnewline
\textbf{SigLIP2} & 1.14B & - & google/siglip2-so400m-patch14-384 \tabularnewline
\textbf{MoCa} & 3.75B & - & moca-embed/MoCa-Qwen25VL-3B \tabularnewline
\bottomrule
\end{tabular}
}
\par\end{centering}
\caption{\label{tab:hf_ckpt}Summary of backbones, their learning rates used for finetuning and specific Huggingface checkpoints.}

\end{table*}

When possible, we set the batch size to 2048 and the corresponding learning rate for each backbone is reported in Table~\ref{tab:hf_ckpt}. In LF-Amazon-131K, we are forced to reduce the batch size to fit experiments in a single GPU. We, therefore, set it to 1024 for distilBERT and 512 for BERT, reducing learning rates to 1e-4 and 5e-5, respectively. For decoders, we train in this dataset with a batch size of 512 using a learning rate of 2e-5. Additionally, when scaling to the 7B model in LF-AmazonTitles-131K we reduce the batch size to 1024 and keep the learning rate to 5e-5. All the remaining parameters (image projection, free vectors and label prototype network) use a learning rate of 5e-4 in encoder models and 1e-4 in decoders.
Additionally, we use AdamW optimizer and adopt hyperparmeters in PRIME~\cite{Dahiya25}. Note that PRIME paper does not use MM-AmazonTitles-300K, dataset where we use the same parametrization as in LF-AmazonTitles-131K. ViXML based on PRIME method can be trained with DistilBERT backbone in around 4.3h (excluding the time to extract image embeddings) for LF-AmazonTitles-1.3M, while 
the same approach with dual-decoder learning based on Qwen2.5-3B-Instruct backbone takes 178h to complete. Please, note that we use right padding across all datasets and models during tokenization, restricting the maximum sequence length to 128 for LF-Amazon-131K and 32 in the remaining datasets.
Another detail that might be of interest is how we treat positional IDs, where we simply assign the corresponding token position in the final prompt, regardless of being a text or an image embedding. Then, if three images are provided, they have three different tokens and position IDs.

\section{Image count impact}
\label{app:num_images}
Table~\ref{tab:results_num_imgs} presents the impact of sweeping the number of images exploited in ViXML on MM-AmazonTitles-300K dataset. Adding a single image yields a +2\% in P@1, while each additional image provides incremental improvements. These results reveal that while increasing image count consistently enhances performance, the most significant gain occurs when transitioning from text-only to incorporating just one image.
\input{results_num_images}

\section{Alternative prompting}
\label{app:alternative_prompts}
In Section 4.4, our goal was to highlight the positive impact of introducing a vanilla template on performance across both uni-modal and multi-modal settings. These prefixes do not carry meaningful semantic content; rather, they serve to lightly contextualize the input, signaling to the LLM that a sentence is being provided. We experimented with variations such as “\textit{This sentence: [text] means in one word:}", “\textit{This sentence text [text]}" and other similar formulations. All of them yielded comparable results (within 0.1 P@1 variation) as presented in Table~\ref{tab:results_alternative_prompts}. This suggests that, at least for short templates, the semantic content of the vanilla prefix is not a critical factor. We, therefore, hypothesize that the performance gains stem more from the structural cue provided by the template than from its semantic richness. In contrast, incorporating meaningful semantic contexts, such as product-related information from retrieval-augmented generation (RAG) systems, could potentially enhance the representation of short queries or labels. In Table~\ref{tab:results_alternative_prompts} we report an example of performance with last token pooling, which slightly degrades performance.
\input{results_alternative_prompts}

\section{Seed variability}
\label{app:seed_variability}
In Table~\ref{tab:results_seeds} we present an example of performance variations for several seeds, exhibiting a low variability across metrics.

\begin{table}[t!]
      \centering
      \resizebox{0.55\linewidth}{!}{
        \begin{tabular}{@{}l|l|cccc@{}}
        \toprule
        \textbf{Method} & \textbf{XMC} & \textbf{P@1} & \textbf{P@5} & \textbf{PSP@1} & \textbf{R@100}  \\ \midrule
        \multirow{5}{*}{\textbf{Qwen2.5-0.5B-I}}
        & \textbf{Seed 1} & 44.84 & 21.66 & 39.64 & 70.30  \\
        & \textbf{Seed 2} & 44.84 & 21.72 & 39.67 & 70.47  \\
        & \textbf{Seed 3} & 44.71 & 21.64 & 39.54 & 70.24  \\
        & \textbf{Seed 4} & 45.03 & 21.68 & 39.77 & 70.39  \\
        & \textbf{Seed 5} & 44.88 & 21.68 & 39.72 & 70.38  \\
        
        \bottomrule
    \end{tabular}
    }
    \caption{Seed variability for dual-decoder learning in LF-AmazonTitles-131K.} \label{tab:results_seeds}
\end{table}

\section{Retrieval-augmented inference}
\label{app:rae}
Related literature commonly employs pipelines that enhance embedding-based predictions~\cite{Dahiya23,Kharbanda24,Prabhu25}. It is widely adopted the use of extreme classifiers, while recently authors in~\cite{Wang25} propose a retrieval augmented encoders inference strategy that exploits training data samples and their positive labels ground-truth to enhance predictions. Interestingly, the authors present in their paper remarkable performance improvements when conducting this inference with relatively weak embeddings.
We are, therefore, interested in analyzing the potential of pairing this retrieval augmented inference with our proposals, which represent a much better starting point than that of~\cite{Wang25} work.
In Table~\ref{tab:results_rae}, we present these results demonstrating that retrieval augmented inference also boosts performance when paired with much better base methods than those in~\cite{Wang25}. Specially, in LF-AmazonTitles-1.3M where there is a richer training ground-truth, we boost P@1 over 2 points in most configurations. It is worth mentioning that propensity metric (PSP@1) is negatively impacted by this inference, as leveraging ground-truth information promotes its inherent bias towards head labels.
We implement this inference strategy as follows:
\input{results_rae}

\begin{itemize}
    \item We conduct two separate searches: firstly on the label embeddings and secondly on the training data query embeddings. This is different from~\cite{Wang25}, where they search on a single extended space (labels and training queries). We introduced this modification because training queries many times double the number of labels, which reduces the importance of labels. From here, we follow the inference strategy as proposed in ~\cite{Wang25}, being the only difference that our scores come from two separate searches rather than a unified search.
    \item Retrieving the closest labels provides label predictions, which are given a $\lambda$ weight. For the second search on training queries, the scores of the retrieved items are weighted by $1-\lambda$.
    \item Both results are unified to mimic a single search and are subsequently softmax-normalized for each query (temperature of 0.05). Then, training queries are turned into label predictions by selecting their relevant labels as defined in the ground-truth.
    \item After this conversion, refined scores with the support of the retrieval process on an augmented search space are available. Note that we constrain the search on both spaces to 100 items and use $\lambda=0.9$ to mainly rely on the base method predictions. Authors in~\cite{Wang25} use $\lambda=0.5$ given that, as opposed to our strong base method, they use a poorly performing model that greatly benefits from higher weights for the training queries contribution.
    
\end{itemize}

\section{Extended comparison}
\label{app:extended_comp}
In this section we extend the comparison against related literature (see Table~\ref{tab:results_comparison_extended}) aiming at highlighting the relevant improvements we achieve against all available literature. We keep the same methods as in the main paper and add the following classifier methods: XR-Transformer \cite{Zhang21}, LightXML~\cite{Jiang21}, ELIAS~\cite{Gupta22}, CascadeXML~\cite{Kharbanda22}. We further add the vanilla dual-encoder method NGAME~\cite{Dahiya23} for completeness. Our dual-decoder learning proposal and the multi-modal ViXML framework introduced in this paper substantially outperform all related literature.

\begin{table}[t!]
\begin{centering}
\resizebox{0.72\columnwidth}{!}{
    \begin{tabular} {@{}l|cc@{\hspace{1em}}|ccc|ccc}
        \toprule
        \textbf{Method} & \textbf{E} & \textbf{M} & \multicolumn{3}{c|}{\textbf{~~LF-AmazonTitles-131K~~}} & \multicolumn{3}{c}{\textbf{LF-Amazon-131K}} \\
        \midrule
        & & & \textbf{P@1} & \textbf{P@5} & \textbf{PSP@1} & \textbf{P@1} & \textbf{P@5} & \textbf{PSP@1} \\
        \midrule
       \textbf{LigthXML} & \cmark & \xmark & 35.60 & 17.45 & 25.67 & 41.49 & 20.75 & 30.27 \\
        \textbf{CascadeXML} & \cmark & \xmark & 35.96 & 18.15 & - & - & - & - \\
        \textbf{XR-Transf.} & \cmark & \xmark & 38.10 & 18.32 & 28.86 & 45.61 & 22.32 & 34.93 \\
        \textbf{ELIAS} & \cmark & \xmark & 40.13 & 19.54 & 31.05 & - & - & - \\
        \textbf{DEXML} & \xmark & \xmark & 42.52 & 20.64 & - & - & - & - \\
        \textbf{UniDEC} & \cmark & \xmark & 44.35 & 21.03 & 39.23 & 47.80 & 23.35 & 40.28 \\
        \textbf{PRIME} & \cmark & \xmark & 45.26 & 21.48 & 39.29 & 48.20 & 23.28 & 40.16 \\
        \textbf{NGAME} & \cmark & \xmark & 46.01 & 21.47 & 38.81 & 46.65 & 22.03 & 38.67 \\
        \textbf{Renée} & \cmark & \xmark & 46.05 & 22.04 & 39.08 & 48.05 & 23.26 & 40.11 \\
        \textbf{DEXA} & \cmark & \xmark & 46.42 & 21.59 & 39.11 & 47.16 & 22.42 & 38.70 \\
        \textbf{OAK} & \cmark & \cmark & 46.42 & 21.88 & 39.76 & - & - & - \\
        \textbf{MOGIC} & \cmark & \cmark & 47.01 & 22.40 & 40.62 & - & - & - \\
        \textbf{PINA} & \cmark & \xmark & - & - & - & 46.76 & 23.20 & - \\
        \midrule
        \textbf{ViXML (Ours)} & \cmark & \cmark & \textbf{53.08} & \textbf{25.74} & \textbf{46.22} & \textbf{55.57} & \textbf{26.84} & \textbf{47.41} \\
        \midrule
        \midrule
        \textbf{Method} & \textbf{E} & \textbf{M} & \multicolumn{3}{c|}{\textbf{MM-AmazonTitles-300K}} & \multicolumn{3}{c}{\textbf{LF-AmazonTitles-1.3M}} \\
        \midrule
        & & & \textbf{P@1} & \textbf{P@5} & \textbf{PSP@1} & \textbf{P@1} & \textbf{P@5} & \textbf{PSP@1} \\
        \midrule
        \textbf{MUFIN} & \cmark & \cmark & 52.30 & 34.76 & - & - & - & - \\
        \textbf{CascadeXML} & \cmark & \xmark & - & - & - & 47.82 & 38.31 & 17.17 \\
        \textbf{OAK} & \cmark & \cmark & - & - & - & 49.46 & 38.61 & 34.92 \\
        \textbf{XR-Transf.} & \cmark & \xmark & - & - & - & 50.14 & 39.98 & 20.06 \\
        \textbf{MOGIC} & \cmark & \cmark & - & - & - & 50.95 & 39.95 & \textbf{36.28} \\
        \textbf{Renée} & \cmark & \xmark & - & - & - & 56.04 & 45.32 & 28.54 \\
        \textbf{DEXA} & \cmark & \xmark & - & - & - & 56.63 & 43.90 & 29.12 \\
        \textbf{NGAME} & \cmark & \xmark & - & - & - & 56.75 & 44.09 & 29.18 \\
        \textbf{UniDEC} & \cmark & \xmark & - & - & - & 57.41 & 45.89 & 30.10 \\
        \textbf{DEXML} & \xmark & \xmark & - & - & - & 58.40 & 45.46 & 31.36 \\
        \textbf{PRIME} & \cmark & \xmark & - & - & - & 59.62 & 46.75 & 31.20 \\
        \midrule
        \textbf{ViXML (Ours)} & \cmark & \cmark & \textbf{57.37} & \textbf{38.08} & \textbf{36.12} & \textbf{67.83} & \textbf{53.72} & 35.40 \\
        \bottomrule 
    \end{tabular}
}
\par\end{centering}
\caption{Extended comparison against related work. We report best numbers available for all works, regardless of using some inference enhancement (E) or meta-data (M).}
\label{tab:results_comparison_extended}
\end{table}

\section{Latencies}
\label{app:latency}
ViXML operates at inference time following efficient common retrieval practices, \ie maximum inner product search using embeddings. In particular, most time is spent encoding inputs: 
a BERT model encodes text-only inputs of length 32 in 0.034 ms, while Qwen2.5-3B-Instruct takes 1.11 ms; when adding images BERT-based ViXML takes 0.037 ms, while ViXML based on Qwen2.5-3B-Instruct spends 1.27 ms. Each query must be encoded at inference time, whereas label embeddings are precomputed. For images, encoding with the 1.14B SigLIP2 model at 384×384 resolution takes 4.28 ms per image. Time retrieving top-k labels via Approximate Nearest Neighbor search is negligible.
Importantly, the reported latencies reflect an unoptimized implementation. In practice, standard techniques such as quantization, KV caching, and efficient inference engines like VLLM or DeepSpeed-Inference could significantly reduce latency.

\section{Limitations}
\label{app:limitations}

Our work achieves substantial improvements in XMC performance through dual-decoder learning and visual metadata integration with ViXML. However, several limitations require discussion and suggest avenues for future work. While dual-decoder learning maintains manageable computational overhead, scaling to LLM backbones inevitably increases training times, irrespective of visual metadata usage. This computational challenge opens several compelling research directions:

\begin{itemize}
    \item \textbf{Efficiently scaling to larger LLMs:} While larger language models could potentially boost performance, they would require prohibitively long training times. Developing more efficient optimization strategies for faster convergence to optimal performance could help in overcoming this challenge.
    
    \item \textbf{Visual representations:} ViXML uses a single embedding per image, which might be suboptimal for capturing fine-grained visual information. However, this approach stands as a tradeoff between efficiency (slight extension of text model length) and performance (ViXML clearly boosts text-only alternatives).
    Finetuning the vision encoder or leveraging token-level image embeddings could potentially yield better performance at the cost of additional memory requirements and increased training times.
    
    \item \textbf{Vision-language model integration:} A promising direction involves finetuning vision-language models such as Qwen2.5VL, which are pre-trained to jointly process text and visual modalities. However, these models typically generate long sequences of image tokens, creating substantial efficiency trade-offs that require careful consideration.
    
    \item \textbf{Advanced prompt engineering:} Given the crucial role of prompt design in LLM performance, our current prompting strategies might be suboptimal. A more systematic exploration of prompt engineering techniques could yield significant performance improvements with minimal computational overhead.

    \item \textbf{Absence of images:} ViXML is designed under the assumption that visual metadata is mostly available and we did not explore other scenarios. However, we know that the performance of models trained with image and text lies under the performance of text-only models when images are removed. This highlights an important limitation to address in future work: preparing the model to deal with image sparsity and even XMC scenarios where queries are text-only and labels are multi-modal (and vice-versa).
    \item \textbf{Larger datasets:} We are currently using the largest publicly available dataset, LF-AmazonTitles-1.3M. However, it is reasonable to think of larger search spaces. From an embedding perspective, further scaling would only require playing with embedding dimensionality and/or Approximate Nearest Neighbor strategies. The biggest challenge would be the training time when using decoder-only LLMs.  
\end{itemize}

Beyond these technical considerations, we identify broader limitations that extend throughout the XMC literature. First, the embeddings learned in XMC are inherently task-specific, limiting their broader applicability. Integrating XMC capabilities into general-purpose text or multi-modal embedding frameworks could produce unified models capable of addressing multiple tasks simultaneously, representing a significant advancement for the field. Second, this work does not address zero-shot XMC scenarios, as we assume shared label spaces between training and test phases.


%% file: results_num_images.tex
\begin{table}[t]
      \centering
      \resizebox{0.35\linewidth}{!}{
        \begin{tabular}{@{}l|cccc@{}}
        \toprule
        \textbf{\#I} & \textbf{P@1} & \textbf{P@5} & \textbf{PSP@1} & \textbf{R@100}  \\ 
        \midrule
        0 & 52.40 & 34.29 & 36.36 & 72.59 \\
        \midrule
        1 & 54.51 & 35.57 & 37.14 & 75.64 \\
        2 & 54.80 & 35.76 & 37.35 & 75.97 \\ 
        3 & 55.03 & \textbf{35.91} & 37.52 & 76.07 \\ 
        5 & \textbf{55.10} & 35.88 & \textbf{37.61} & \textbf{76.08} \\ 
        
        \bottomrule
    \end{tabular}
    }
    \caption{Effect of the number of images (\#I) in ViXML.}\label{tab:results_num_imgs}
\end{table}

%% file: results_alternative_prompts.tex

\begin{table}[t]
      \centering
      \resizebox{0.75\linewidth}{!}{
        \begin{tabular}{@{}l|cccc@{}}
        \toprule
        \textbf{Prompts} & \textbf{P@1} & \textbf{P@5} & \textbf{PSP@1} & \textbf{R@100}  \\ \midrule
        “\textit{This product text [text]}" & 47.42 & 22.89 & 41.88 & 74.34  \\
        “\textit{This sentence text [text]}" & 47.39 & 22.86 & 41.89 & 74.22  \\
        “\textit{This product text [text] means in one word:}" & 47.45 & 22.91 & 42.06 & 74.04  \\
        “\textit{This product text [text] means in one word:}" (*) & 47.07 & 22.68 & 41.86 & 73.20  \\
        “\textit{The description of a product is: [text]}" & 47.44 & 22.87 & 42.00 & 74.35  \\

        \bottomrule
    \end{tabular}
    }
    \caption{Prompt variations for text-only dual-decoder learning in LF-AmazonTitles-131K with Qwen2.5-3B-Instruct. All alternatives use average pooling to build sentence embeddings except for (*), where we use last token pooling. End-of-sequence token is always used.} \label{tab:results_alternative_prompts}
\end{table}

%% file: results_rae.tex
\begin{table}[t!]
\begin{centering}
\resizebox{0.80\columnwidth}{!}{
    \begin{tabular} {l|cccc|cccc}
        \toprule
        \textbf{Method} & \multicolumn{4}{c|}{\textbf{~~LF-AmazonTitles-131K~~}} & \multicolumn{4}{c}{\textbf{LF-Amazon-131K}} \\
        \midrule
        & \textbf{P@1} & \textbf{P@5} & \textbf{PSP@1} & \textbf{R@100} & \textbf{P@1} & \textbf{P@5} & \textbf{PSP@1} & \textbf{R@100} \\
        \midrule
        \textbf{DEL} & 44.86 & 21.45 & 39.62 & 68.05 & 47.98 & 23.21 & 40.99 & 76.22 \\
        \textbf{DEL+R} & 45.29 & 21.89 & 39.24 & 69.40 & 48.46 & 23.74 & 40.77 & 77.67 \\
        \textbf{ViXML (DEL)} & 49.55 & 23.73 & 43.78 & 76.73 & 51.30 & 24.69 & 44.09 & 81.40 \\
        \textbf{ViXML (DDL + R)} & 50.02 & 24.24 & 43.47 & 77.89 & 51.85 & 25.20 & 43.94 & 82.56 \\
        \textbf{DDL} & 48.06 & 23.07 & 42.43 & 75.24 & 53.17 & 25.72 & 45.74 & 85.61 \\
        \textbf{DDL+R} & 48.41 & 23.39 & 41.69 & 75.89 & 53.68 & 26.14 & 45.53 & 86.17 \\
        \textbf{ViXML (DDL)} &  52.75 & 25.37 & \textbf{46.70} & 83.27 & 55.11 & 26.46 & \textbf{47.73} & 87.73 \\
        \textbf{ViXML (DDL+R)} & \textbf{53.08} & \textbf{25.74} & 46.22 & \textbf{83.88} & \textbf{55.57} & \textbf{26.84} & 47.41 & \textbf{88.22} \\
        \midrule
        \midrule
        \textbf{Method} & \multicolumn{4}{c|}{\textbf{MM-AmazonTitles-300K}} & \multicolumn{4}{c}{\textbf{LF-AmazonTitles-1.3M}} \\
        \midrule
        & \textbf{P@1} & \textbf{P@5} & \textbf{PSP@1} & \textbf{R@100} & \textbf{P@1} & \textbf{P@5} & \textbf{PSP@1} & \textbf{R@100} \\
        \midrule
        \textbf{DEL} & 52.40 & 34.29 & 36.36 & 72.59 & 58.49 & 45.27 & 33.28 & 63.76 \\
        \textbf{DEL+R} & 53.94 & 36.16 & 34.45 & 76.79 & 60.98 & 48.18 & 28.99 & 68.16 \\
        \textbf{ViXML (DDL)} & 55.03 & 35.91 & 37.52 & 76.07 & 64.17 & 49.43 & 36.58 & 70.65 \\
        \textbf{ViXML (DEL+R)} & 56.37 & 37.54 & 35.65 & 79.50 & 66.53 & 52.36 & 33.09 & 75.08 \\
        \textbf{DDL} & 54.71 & 35.93 & 36.29 & 77.22 & 60.74 & 47.39 & 34.01 & 66.88 \\
        \textbf{DDL+R} & 55.48 & 37.00 & 34.74 & 78.93 & 62.84 & 49.76 & 31.11 & 70.24 \\
        \textbf{ViXML (DDL)} & 56.55 & 37.04 & \textbf{37.60} & 79.38 & 66.01 & 51.21 & \textbf{37.71} & 73.13 \\
        \textbf{ViXML (DDL+R)} & \textbf{57.37} & \textbf{38.08} & 36.12 & \textbf{80.85} & \textbf{67.83} & \textbf{53.72} & 35.40 & \textbf{76.76} \\
        \bottomrule 
    \end{tabular}
}
\par\end{centering}
\caption{Retrieval Augmented (R) inference effect when paired with our dual-encoder learning (DEL), dual-decoder learning (DDL) and our multi-modal framework ViXML. For DEL approaches we report results for DistilBERT model.}
\label{tab:results_rae}
\end{table}